\documentclass[10pt,twocolumn,letterpaper]{article}

\usepackage{wacv}              

\usepackage{graphicx}
\usepackage{amsmath}
\usepackage{amssymb}
\usepackage{booktabs}
\usepackage{balance}
\usepackage[accsupp]{axessibility} 

\usepackage[pagebackref,breaklinks,colorlinks]{hyperref}

\usepackage[capitalize]{cleveref}
\crefname{section}{Sec.}{Secs.}
\Crefname{section}{Section}{Sections}
\Crefname{table}{Table}{Tables}
\crefname{table}{Tab.}{Tabs.}

\def\wacvPaperID{08} 
\def\confName{WACV}
\def\confYear{2025}

\begin{document}

\title{Iris Recognition for Infants}

\author{Rasel Ahmed Bhuiyan$^*$, Mateusz Trokielewicz$^{\dag,\ddag}$, Piotr Maciejewicz$^{**}$, Sherri Bucher$^{***}$, Adam Czajka$^*$\\
$^*$University of Notre Dame, IN, USA, $^{\dag}$Warsaw University of Technology, Poland, $^{\ddag}$PayEye Poland,\\$^{**}$Medical University of Warsaw, Poland, $^{***}$Indiana University Indianapolis, IN, USA\\
{\small Corresponding email: {\tt rbhuiyan@nd.edu}}
}
\maketitle

\begin{abstract}

Non-invasive, efficient, physical token-less, accurate and stable identification methods for newborns may prevent baby swapping at birth, limit baby abductions and improve post-natal health monitoring across geographies, within the context of both the formal (i.e., hospitals) and informal (i.e., humanitarian and fragile settings) health sectors. This paper explores the feasibility of application iris recognition to build biometric identifiers for 4-6 week old infants. We (a) collected near infrared (NIR) iris images from 17 infants using a specially-designed NIR iris sensor; (b) evaluated six iris recognition methods to assess readiness of the state-of-the-art iris recognition to be applied to newborns and infants; (c) proposed a new segmentation model that correctly detects iris texture within infants iris images, and coupled it with several iris texture encoding approaches to offer, to the first of our knowledge, a fully-operational infant iris recognition system; and, (d) trained a StyleGAN-based model to synthesize iris images mimicking samples acquired from infants to deliver to the research community privacy-safe infant iris images. The proposed system, incorporating the specially-designed iris sensor and segmenter, and applied to the collected infant iris samples, achieved Equal Error Rate (EER) of 3\% and Area Under ROC Curve (AUC) of 99\%, compared to EER$\geq$20\% and AUC$\leq$88\% obtained for state of the art adult iris recognition systems. This suggests that it may be feasible to design methods that succesfully extract biometric features from infant irises. 
\end{abstract}

\section{Introduction}
\subsection{Background and Motivation}


The patient safety of newborns via consistent, reliable, safe, effective, and easy-to-implement non-invasive identification has long been of global interest in order to reduce medical errors and to prevent baby swapping or crimes such as abduction. These risks exist across both high- and low-resource geographies, particularly in hospitals without the ability to deploy sophisticated patient identification security protocols. This includes settings where the potential for newborn misidentification is high, namely (a) due to similar birth dates, medical record numbers, or common surnames, (b) in the case of multiple gestation births (twins and triplets), or (c) within the Newborn Unit (NBU) or Neonatal Intensive Care Unit (NICU) \cite{Gray_Ped_2006}. Hospitals that lack effective identity management systems are more prone to deleterious patient safety and medical error  incidents, which can have severe consequences, including emotional distress for families, legal battles to establish biological parentage, and significant damage to a hospital’s reputation and finances. 

As an example, each year approx. 20,000 babies are switched due to various forms of newborn misidentification \cite{segal2018twins}. A study conducted at Boston’s Beth Israel Deaconess Medical Center found that around 26\% of neonates in the NICU were at risk of misidentification due to similar identifiers \cite{gliklich2014registries}. In a study involving health professionals from 54 hospitals in the Vermont Oxford Network, it was found that 11\% of newborns over two years were misidentified \cite{chassin2002wrong}. A study by the National Center for Missing and Exploited Children (NCMEC) reveals alarming statistics on infant abductions \cite{NCMEC_2021}: from 1997 to 2016, 66\% of abducted infants were taken from hospitals, with the majority being less than six months old. Among those abductions, 58.6\% happened in the mother’s room, 13.6\% in the nursery, 12.1\% in pediatrics, and 15.7\% elsewhere on the premises.
Although Gaille \cite{gailleRareBabies2017} notes that many of these issues are resolved before families leave the hospital, the risk of baby switching could have been lowered by application of non-invasive, fast and affordable biometric identification means. Additionally, if such a method would offer reliable identification throughout the life, without the need of re-enrolling the subject, this could also support prevention of newborn abductions and help in searching for missing children, as well as, potentially, facilitate linkage of health records across the life course \cite{Freytsis_FinB_2021}. Another area of growing concern for improved methods of reliable newborn identification is within fragile and humanitarian settings, such as refugee camps. Implementing methods by which to improve the stable identification of infants within these dynamic settings may not only improve safety for vulnerable populations as they navigate often chaotic and unfamiliar systems, with an elevated risk of family separation, but also may lead to more effective monitoring and evaluation of infant health, and linkage with routine health information systems at the patient and population levels \cite{Tappis_BMC_2021}. Iris recognition is considered to be accurate and relatively stable over time \cite{Daugman_PAMI_1993}, keeping its identification capabilities even shortly after death \cite{Trokielewicz_BTAS_2016,Sansola_MastersThesis_2015}. (Although a hypothesis about the iris pattern long-term stability has still to be confirmed \cite{Trokielewicz_BF_2015,Bowyer_IET_2015} and has not yet been documented for cases when the enrollment is done during infancy.) Iris recognition sensors and algorithms designed for adults were shown to be effective for children aged two and older, but they are not applicable for infants under two years old \cite{nelufule2023infant}.

\subsection{Novel Contributions}
This paper explores the feasibility of iris recognition applied to infants of age between 4 and 6 weeks, and -- to our knowledge for the first time -- proposes the iris recognition method designed specifically for infants. The \textbf{key contributions} of this work are:

\begin{itemize}
    \item[(c1)] collection of the unique dataset consisting of 1,920 iris images acquired from 17 newborns at the neonatology clinic with a custom-designed (specifically for this study) iris sensor producing ISO/IEC 19794-6-compliant \cite{ISO_29794-6_2016} iris images;

    \item[(c2)] an iris segmentation model capable of segmenting correctly iris images acquired from newborns (in addition to being capable of segmenting other types of iris samples, such as those with eye diseases or forensic samples acquired after death); this segmenter was integrated with an iris encoding routines and custom-designed iris sensor to create a fully-operational, first infant iris recognition system;

    \item[(c3)] newborn iris recognition experiments using six state-of-the-art iris recognition methods, each implementing a different approach to iris recognition, including classical Gabor wavelets-based, human saliency-based, deep learning-based, and  commercial methods;

    \item[(c4)] to protect children privacy, and to meet data collection restrictions, instead of publishing collected newborn iris images (c1), with this paper we offer a dataset of 1,000 synthetic newborn iris samples generated by a StyleGAN-based, identity leakage-free generative model trained on authentic newborn iris images (c1).

\end{itemize}


\subsection{Prenatal and Early Postnatal Iris Development}
Embryonic development of the iris begins at the 12th week of pregnancy \cite{sunita2021anatomical, sturm2009genetics, moazed2020iris}. The unique surface patterns of the iris result from the individual arrangement of Fuchs' crypts and contraction furrows, which are stromal defects caused by embryonic regression of the mesoderm during eye anterior chamber formation. Development of crypts on the iris surface starts around the 20th week of pregnancy and  probably ends in the postnatal period. As the diameter of the cornea increases, the diameter of the iris and the width of the pupil increase too. Other postnatal changes to the eye involve the anatomical elements around the iris. Among other things, the thickness of the cornea decreases, the axial length of the eyeball increases, and the depth of the anterior chamber adjacent to the iris increases. A frequently visible postnatal change in the iris is a change in iris color. This change will occur within the first 6–12 months of life due to the accumulation of pigment in iris stromal melanocytes. Eye color is a polygenic trait influenced by as many as 16 genes, with HERC2 and OCA2 on chromosome 15 playing the most significant roles. Uneven pigment distribution can cause iris heterochromia, which does not impact automatic iris recognition due to near infrared iris scanning. One important difference between adult and infant iris scans, not related to anatomy, though, is usually a larger dilation of the pupil in case of newborns. This happens when such scans are taken as part of the routine screening for retinopathy, which requires administering mydriatic agents.

\subsection{Small Data Statement} 
{\bf Qualification as small data research:} This paper focuses solely on infants of age between 4 and 6 weeks, and proposes the first iris recognition method designed specifically for infants with a very limited data. {\bf Methods to tackle the data size and privacy challenges:} (a) we have applied ophthalmology-driven infant iris image augmentations simulating excessive near infrared light reflections from the retina applied to adult iris images to make the infant-specific segmentation model's training effective; and (b) we trained a generative model to synthesize newborn iris images without leaking identity of infants, what allows to increase the training data sample without compromising privacy of kids.

\section{Related Works}
Infant recognition has garnered increasing attention in recent years for various applications. Several studies have explored different biometric modalities for this purpose. Bharadwaj \etal \cite{bharadwaj2016domain} proposed an autoencoder-based approach for newborn face recognition, achieving a rank-1 identification accuracy of 78.5\% and a verification accuracy of 63.4\% at a 0.1\% false accept rate . Liu \cite{liu2017infant} investigated infant recognition through footprint analysis, introducing a minutia descriptor based on deep convolutional neural networks, utilizing a dataset of 60 subjects aged 1 to 9 months. Other studies have employed multimodal biometric traits, such as face, fingerprint, and ear recognition, to identify newborns \cite{moolla2021biometric, basak2017multimodal}. However, a significant challenge with these traits is that they are often not permanent and change rapidly in the first months of life. In contrast, iris 
patterns are hypothesized to be fully formed by the eighth month of the gestation \cite{daugman2007new}, making iris an appealing biometric trait for infants compared to other modalities.

Several studies have investigated iris recognition for young children, revealing both challenges and advancements. Johnson \etal \cite{johnson2018longitudinal} examined iris recognition in children aged 4 to 12 years over multiple visits using the VeriEye matcher \cite{verieye}. They found a slight, statistically insignificant decrease in match scores over 12 months, suggesting stability in iris characteristics from age of 4 years onwards.

Hutchison \cite{hutchison2018longitudinal} conducted a longitudinal study on iris recognition in infants aged 0 to 2 years. This study compared iris image quality metrics between infants and adults, assessing performance across age groups: 0-6 months, 7-12 months, 13-24 months, and adults. The analysis revealed that while image quality varied, infants aged 13-24 months had iris images more similar (in quality terms) to adults. However, infants aged 0-6 months showed poorer performance at a false match rate of 0.01\% compared to older infants and adults.

Nelufule \etal \cite{nelufule2019image} explored image quality assessment methods for iris biometrics in minors. They applied various quality assessment techniques (light variation, pupil dilation, off-angle capture assessment, and pixel count) to children's iris images and compared them with adult images from the CASIA iris database. Their findings indicated that, after removing images with no visible iris area, the iris image quality metric distributions for children’s iris images were comparable to those of adults. Building on this, Nelufule \etal \cite{nelufule2020circular} developed an iris detection method for infants using circular Hough transform. 

Das \etal (2021) \cite{das2021iris} conducted a longitudinal study of iris recognition among 209 children aged 4 to 11 years over three years. They found a statistically significant aging effect, but it was minor compared to other variability factors. Iris recognition remained effective for up to three years between samples, despite challenges with enrolling very young children. This study contributed a unique dataset of longitudinal iris images for this age group.

Moolla \etal (2021) \cite{moolla2021biometric} examined biometric recognition systems for infants, focusing on fingerprint, iris, and outer ear shape biometrics. Their research highlighted that preprocessing adjustments improved the localization and segmentation of infant irises, with successful matching starting from as early as six weeks and improving with age.

Recent work by Nelufule \etal (2023) \cite{nelufule2023infant} focused on using infant iris biometrics to identify newborns and young children. They collected iris images using an IriShield-USB BK 2121U camera and evaluated image quality before segmentation. The results demonstrated effective recognition for children aged two and older, but less effective for those under two years old.

{\bf This paper differs from, and extends previous studies by several novel components}. First, it's focused solely on infants. Secondly, it uses a specially-designed for this work newborn iris sensor instead of commercial-off-the-shelf scanner designed for adults, what increases significantly chances of acquiring good-quality NIR iris images in the NICU environment. Thirdly, it proposes a deep learning-based iris segmenter that not only detects the iris in infant iris images, but performs the full iris segmentation in both newborn and adult samples, including forensic post-mortem iris images. Fourthly, this study includes designing a StyleGAN-based generative model and offers 1,000 synthetic (thus, addressing the newborn's privacy) iris images mimicking authentic iris pictures taken from infants.

\section{Infant Iris Data Collection}
\subsection{Data Acquisition Protocol}


The study and imaging of the children's eyes were approved by the Bioethical Committee of the Medical University of Warsaw, Poland (approval doc. number KB/108/2023). Iris photographs were taken at the neonatology clinic at the above university. The photos were taken after obtaining informed consent from the children's guardians. To avoid additional stress factors, only children who were previously scheduled for fundus examination were included in the study. Iris images were taken after pharmacological mydriasis, after local drop anesthesia and after placement of a palpebral fissure dilator. All these activities are elements of standard clinical procedures when examining the fundus of the eye with a Fison indirect ophthalmoscope in the first weeks of life. Thus, capturing the iris images did not require any additional medications or preparations beyond the standard clinical procedure, and extended the examination time by approximately 10 seconds. The iris photographs were anonymized, labeled with a sequential number, designation of the left or right eye, the child's age, and a date of the procedure.

\subsection{Data Acquisition Scanner}

\paragraph{Sensor's characteristics} The data collection was carried out with a prototype device, custom-designed, built and provided by an industrial partner for this study. It is capable of very high quality, high resolution imaging of the human iris close-up, revealing its texture in great detail. The device utilizes a single, 4 megapixel CMOS sensor sensitive in the NIR spectrum of 700 nm and above, as well as NIR illuminators with a peak at 810 nm. This setup is able to provide iris images that surpass current commercial iris sensors offering images with iris representation spanning approximately 900 pixels across the iris diameter, \ie five times more than recommended by ISO/IEC 19794-6 to classify an iris image as the highest-quality biometric sample.

\paragraph{Sensor's safety measures} The NIR illuminators applied in this device, OSRAM SFH 4787S, fall into the class exempt group according to the eye safety standard IEC 62471:2006. Due to very vulnerable group of subjects (infants), the authors additionally secured an independent (of the manufacturer) testing of the assembled device for eye safety at a professional testing lab at the Central Institute for Labour Protection -- Research Institute, 16 Czerniakowska Str., Warsaw, Poland. The obtained measurements and PELs (Permissible Exposure Limits) are shown in Tab. \ref{tab:PELs} and allow to conclude that the device is safe for children.

\begin{table}[htb!]
\small
\centering
\caption{Permissible and measured eye safety factors for the infant iris sensor custom-designed for this study.}
\label{tab:PELs}
\begin{tabular}{lcc}
\toprule
& Permissible & {\bf Measured} \\
& Exposure & \\
& Limit (PEL) & \\\midrule
Skin hazard [J/m$^2$] & 35,566 & {\bf 215}\\ 
Cornea and lens hazard [W/m$^2$] & 3,201 & {\bf 22}\\ 
Retina hazard [W/(m$^2$sr)] & 1,404,494 & {\bf 762}\\\bottomrule
\end{tabular}
\end{table}

\begin{figure}[htb!]
    \centering
    \includegraphics[width=\linewidth]{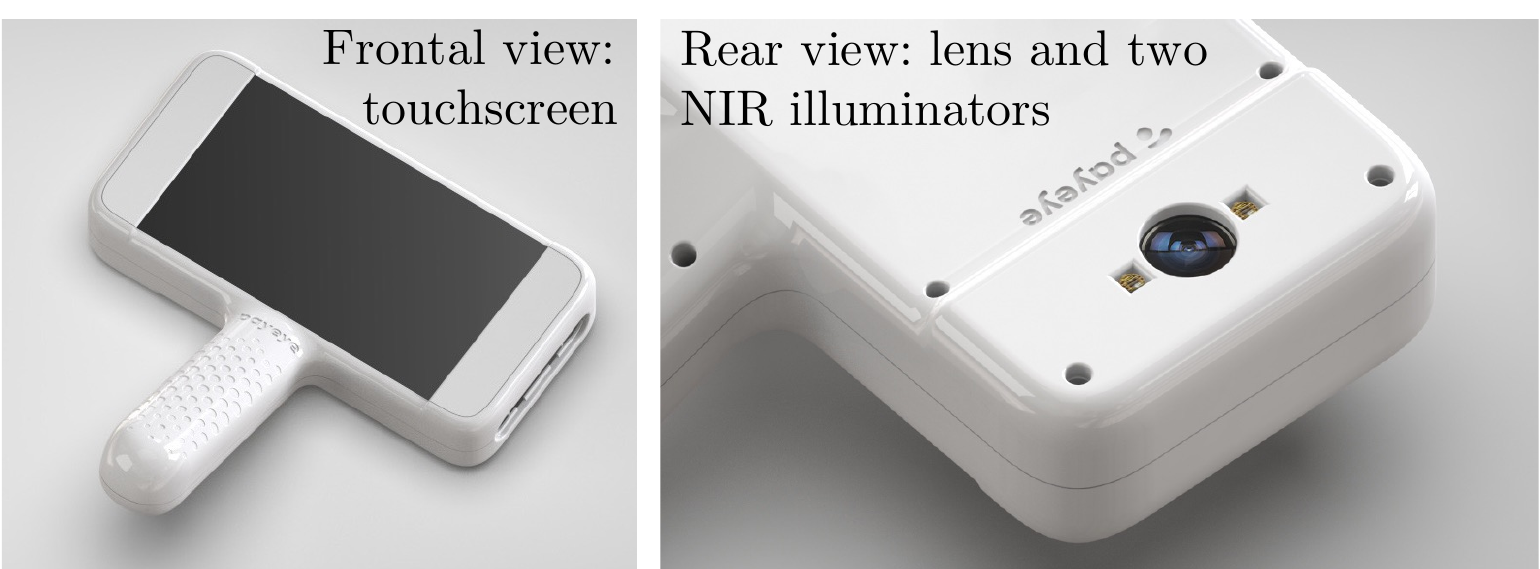}
    \caption[*]{Frontal and rear views of the handheld newborn iris scanner custom-designed for this study by an industrial partner.}
    \label{fig:iris-scanner}
\end{figure}

\subsection{Data Curation}
\label{sec:data-curation}
The initial dataset comprises 4,202 samples acquired from 17 infants aged from 4 to 6 weeks. All samples from a given subject were gathered in a single session in a form of several 10-second sequences. Taking sequences instead of single shots was intentional to maximize collecting at least one good-quality sample per attempt. Due to this procedure, many of the collected images were blurred and had to be discarded. To perform this data curation, we calculated the ISO/IEC 29794-6 \verb+SHARPNESS+ score for each image, which ranges from 0 (low sharpness) to 100 (perfect sharpness). Samples with a \verb + SHARPNESS + score below 10 were omitted, and the final data set used in this study consists of 1,920 total samples from 17 babies.

\subsection{Synthesis of Infant Iris Images}
Due to privacy-related restrictions associated with this study, and to maximally protect vulnerable population of subjects (infants), the original iris images cannot be used for illustrations in this paper, and cannot be disseminated in any form. However, to our best knowledge there are no publicly-available datasets of infant iris images what significantly hampers progress in this research field, primarily due to the challenges associated with data collection, which can only occur in healthcare premises and must be conducted by a medical personnel. To overcome these obstacles, and -- more specifically -- offer illustrations of newborn images and share images mimicking properties of infant iris samples, but at the same time do not reveal identity of children, we trained a StyleGAN2-ADA-based model \cite{karras2020training} with the curated data (as described in Sec. \ref{sec:data-curation}) to synthesize a set of privacy-safe infant iris images.

One of the main challenges in training such a model was the limited amount of data available per subject, which complicates the selection of an appropriate generative model. Traditional models based on Generative Adversarial Networks (GANs) usually require large datasets, but in the case of this study, we have only a few hundred images per subject, leading to potential overfitting of the discriminator. Thus, we chose the StyleGAN2-ADA architecture, which is optimized for training with limited data through various augmentation techniques to enhance the number and diversity of samples. The StyleGAN2-ADA was trained
from scratch using a batch size of 128, with the original iris images rescaled to $256 \times 256$ pixels, and scaled up to $640 \times 480$ after synthesis to match the ISO/IEC 19794-6 recommended resolution.

\begin{figure}[htb!]
    \centering
    \includegraphics[width=\linewidth]{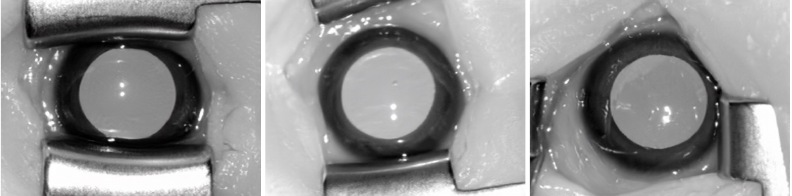}
    \caption{Examples of synthetically-generated infant iris images. The trained generative model offers a remarkable realism of the synthetic images and correctly captured intricate details, such as the iris texture, brighter pupil (compared to adult iris images, in which the pupil is usually darker than iris), eyelid retractors used by medical personnel, specular highlights or skin texture.}
    \label{fig:synthetic-samples}
\end{figure}

As the next step, we utilized the trained model to generate 5,000 synthetic infant iris samples, which then were matched using the HDBIF matcher (see Sec. \ref{sec:matchers}) with all training samples. This matcher calculates the fractional Hamming distance between two binary iris codes. Synthetic samples with a matching score below 0.5 were removed to minimize a potential of a false match between synthesized and authentic samples, and thus to prevent identity leakage. The final synthetic dataset consists of 1,000 samples representing 500 synthetic infant ``identities,'' with two samples per identity. This set of synthetic samples is made publicly available with this work. It is also used to provide illustrations in this paper (Figs. \ref{fig:synthetic-samples} and \ref{fig:segmentation-vis}).

\section{State-of-the-art Iris Matchers}
\label{sec:matchers}

To first assess the feasibility of infant iris recognition with the existing approaches to iris matching, we have employed six state-of-the-art adult iris recognition methods. Five methods are open-source academic solutions, and one is a commercially available product. All methods are shortly described in the following paragraphs.

\paragraph{Human-Driven Binary Image Feature extractor (HDBIF)} is an open-source method that specifically addresses post-mortem iris recognition \cite{czajka2019domain}. It 
combines deep learning-based post-mortem iris-aware segmentation \cite{trokielewicz2020post} with domain-specific feature extraction. 
The comparison score is the fractional Hamming distance between iris codes (as in Daugman's solution \cite{daugman2007new} but with human-driven kernels) representing non-occluded iris portions.

\paragraph{The University of Salzburg Iris Toolkit (USIT) v3.0} is an open-source software for iris recognition \cite{USIT3}. 
The combination of algorithms selected from the USIT toolkit for this work follows the recommendation from the USIT authors, and includes Contrast-Adjusted Hough Transform (CAHT) for iris localization and segmentation \cite{rathgeb2013iris}, the 1D Log-Gabor (LG) filter for feature extraction \cite{masek2003recognition}, and the TripleA algorithm for calculating matching scores.

\paragraph{Open Source for IRIS (OSIRIS) v4.1} is an open-source solution developed under the BioSecure EU project \cite{othman2016osiris}, and directly follows Daugman's methodology \cite{daugman2007new}. The phase quantization of the Gabor filter outcomes is utilized to calculate the iris code, and a comparison score between iris codes is calculated using the fractional Hamming distance. 

\paragraph{Neurotechnology VeriEye} is a commercial iris recognition Software Development Kit \cite{verieye} that implements a proprietary algorithm not yet published (to our knowledge). VeriEye is often ranked as one of the top methods in the NIST IREX program \cite{IREX_X_URL}. For this matcher, a higher similarity score indicates a better match between two iris samples. 
%
%
As recommended by the VeriEye suppliers, scores above 40 indicate a match (genuine pair), while scores at or below 40 indicate no match (impostor pair). 

\paragraph{Dynamic Graph Representation (DGR)} \cite{ren2020dynamic} 
utilizes a hybrid framework that combines convolutional neural networks with graph models to create dynamic graph representations. 
This method forms feature graphs with nodes representing feature vectors and edges indicating node relationships. The resulting graph is processed by SE-GAT, a structure based on Graph Attention Networks (GAT) \cite{velivckovic2017graph}, to further refine the features. 
Notably, the DGR model does not rely on segmentation and has demonstrated superior performance in recognizing occluded biometric data, particularly in iris and face samples.

\paragraph{The WorldCoin Iris Recognition Inference System (WIRIS)} \cite{wldiris} is an open-source iris recognition algorithm, which -- similarly to OSIRIS -- implements Daugman's iris recognition pipeline \cite{Daugman_PAMI_1993}, and consists of four typical for Daugman's algorithm steps: iris image segmentation, normalization, 2D Gabor wavelets-based feature extraction, and Hamming distance-based matching. The segmentation step uses 
an encoder and two decoders: one for estimating eye geometrical parameters and second for detecting occlusions such as eyelashes and hair \cite{lazarski2022two}. 

\section{Infant Iris Segmentation and Recognition}

\subsection{Segmentation Dataset and Infant Iris-Specific Training Data Augmentations}

\begin{figure*}[htb!]
    \centering
    \includegraphics[width=\linewidth]{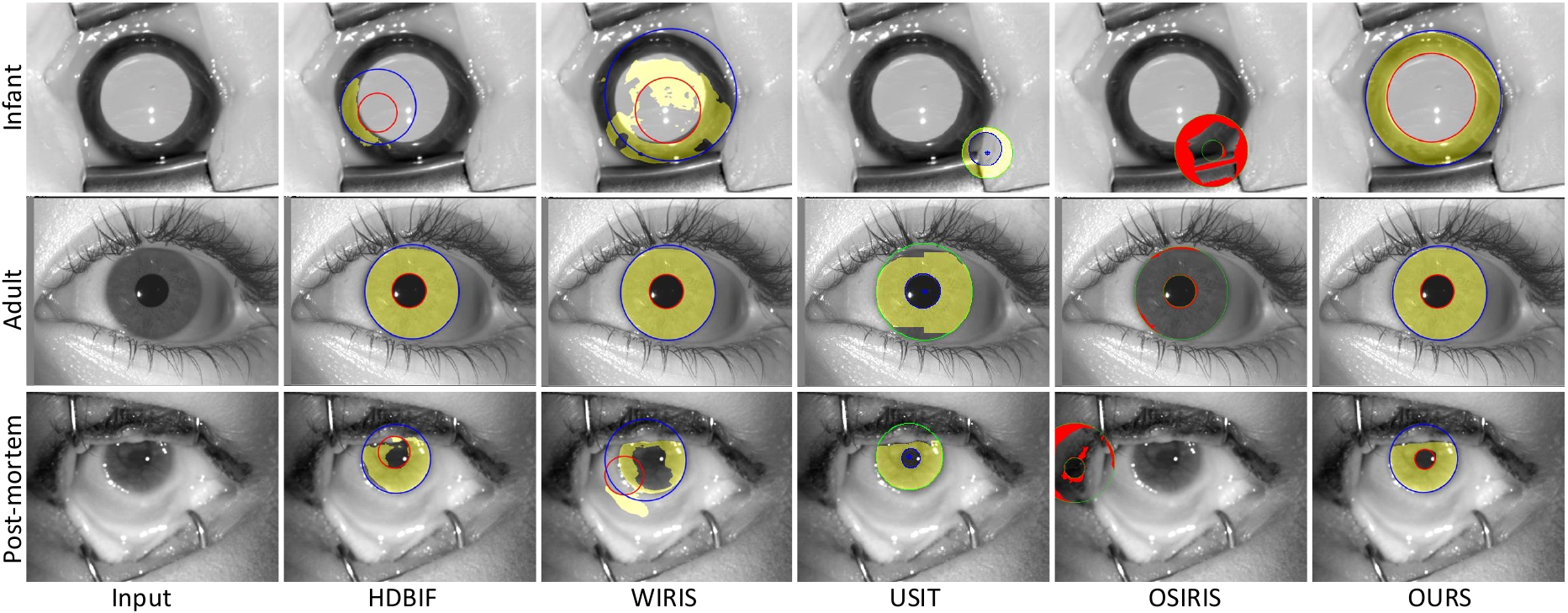}
    \caption{Comparison of iris segmentation visualizations across various states: infant, adult, and post-mortem. The visualizations compare the performance of our developed model with state-of-the-art methods. The model effectively segments irises with varying characteristics, including dark, bright, small, and large pupils.}
    \label{fig:segmentation-vis}
\end{figure*}

As shown in the top row in Fig. \ref{fig:segmentation-vis}, off-the-shelf iris segmentation methods designed for adult eyes are not effective in processing infant iris images. Hypothetically the main challenge for these methods is much brighter pupil area, compared to regular adult iris samples, caused by a reflection of near infrared from retina due to wide opening of iris and illuminators placed closed to the camera optical axis (to minimize the device size). These observations call for designing an infant iris-specific segmentation model.

There are no, to the best of our knowledge, publicly-available datasets of newborn iris images associated with segmentation masks. Thus, we utilized a set of adult iris images with segmentation masks composed of several publicly-available benchmarks (described briefly at the end of this subsection), and applied {\bf infant iris-specific augmentations} to all images to mimic anticipated properties of newborn iris samples. More specifically, for all training adult iris images  
we adjusted the brightness of the iris pupil. To do that, we calculated the minimum and maximum pixel values of the pupil, ranging from 109 to 190, using the circle detection model \cite{ND_OpenSourceIrisRecognition_GitHub} for pupil localization. Then, randomly assigned pixel intensities within this range, and replaced the original iris pupil pixel intensities. Additionally, we applied random rotations between -15 to 15 degrees and Z-normalized the pixel intensity values (to standardize the distribution of pixel intensity to zero mean and unit variance). These proposed augmentations allowed to train a model offering spectacularly good segmentation results, not only for infant iris samples, but also for adult irises, as shown in the last column in Fig. \ref{fig:segmentation-vis}.

The combined training data, including 20,931 adult iris images, was sourced from the following benchmarks: 2,639 images from the CASIA-Iris-Interval-v4 database \cite{CASIA-V4}, 1,283 images from the ND-Iris-0405 database \cite{schott2010frvt}, 800 images from the Warsaw-BioBase-PostMortem-Iris v2.0 \cite{trokielewicz2018iris} and ND-TWINS-2009-2010 \cite{ND-TWINS} (iris images acquired from identical twins) databases, 1,200 images from the BioSec baseline corpus, 2,250 images from the UBIRIS.v2 database \cite{proencca2009ubiris}, which includes visible wavelength images with only the red channel used, and 12,759 images from the OpenEDS dataset \cite{garbin2019openeds}. Each image in this initial set is accompanied by a corresponding ground truth segmentation mask. These ground truth annotations come from the IRISSEG-CC dataset by Halmstad University (for the BioSec database), while the ground truth for the CASIA-Iris-Interval-v4, ND-Iris-0405, and UBIRIS databases has been provided by the IRISSEG-EP dataset by the University of Salzburg. Ground truth masks of the OpenEDS dataset are provided by Facebook. 

\subsection{Proposed Segmentation Model}

In this study, we employed the nested U-Net architecture with dilated convolutions and attention blocks, offered by the open source iris recognition project \cite{ND_OpenSourceIrisRecognition_GitHub}, with a width of 64. That segmentation model is based on a nested U-Net architecture that additionally uses a \texttt{SharedAtrousConv2d} layer implementing (in parallel) standard and dilated convolutions with shared weights. Each block in the network is a \texttt{SharedAtrousResBlock}, comprising two shared atrous convolutions, batch normalization, and ReLU activation, with a residual connection. The nested U-Net structure enables each level to fuse features from previous layers, while bilinear interpolation is used for upsampling and downsampling. The final output layer consolidates features from all nested blocks to produce a segmentation map.

We trained the model from scratch for 300 epochs using a batch size of $32 \times N$, where $N$ is the number of augmentation repetitions. We used the MADGRAD optimizer with a learning rate of 0.001, and a combination of cross-entropy and dice losses. Fig. \ref{fig:segmentation-vis} shows the segmentation results for infant, regular adult and post-mortem iris images for the proposed model and selected state-of-the-art methods. Both the original (sourced from public benchmarks) and augmented (to include infant-specific properties of iris scans) images were used in training. A small set of hand-annotated authentic infant iris images was used in validation to pick the best model by maximizing average intersection over union between predicted and ground truth masks.

\subsection{Infant Iris Recognition Approach}

Infant iris segmentation is the key component that differs from adult iris recognition pipeline. Thus, to create fully-functional infant iris recognition methods, we coupled the proposed new segmenter with a few existing iris feature extraction approaches (HDBIF, OSIRIS, USIT and DGR), which are evaluated in the next section.

\section{Experiments and Results}
\label{sec:results}

\subsection{ISO/IEC 29794-6 Quality of Infant Iris Images}

To evaluate the quality of infant iris images (from the biometric sample quality standpoint), we selected eight ISO/IEC 29794-6 iris image quality metrics and compared their values between infant and adult images. These metrics 
cover factors such as iris pattern visibility, the pupil-to-iris size ratio, pupil shape regularity, gray-scale utilization, and image sharpness. Additionally, we calculated an overall quality metric that integrates these properties, as defined in ISO/IEC 29794-6. The adult samples are sourced from the ND-Iris-0405 database \cite{schott2010frvt}. 

\begin{figure*}[htb!]
    \centering
    \begin{subfigure}{0.25\linewidth}
        \includegraphics[width=\linewidth]{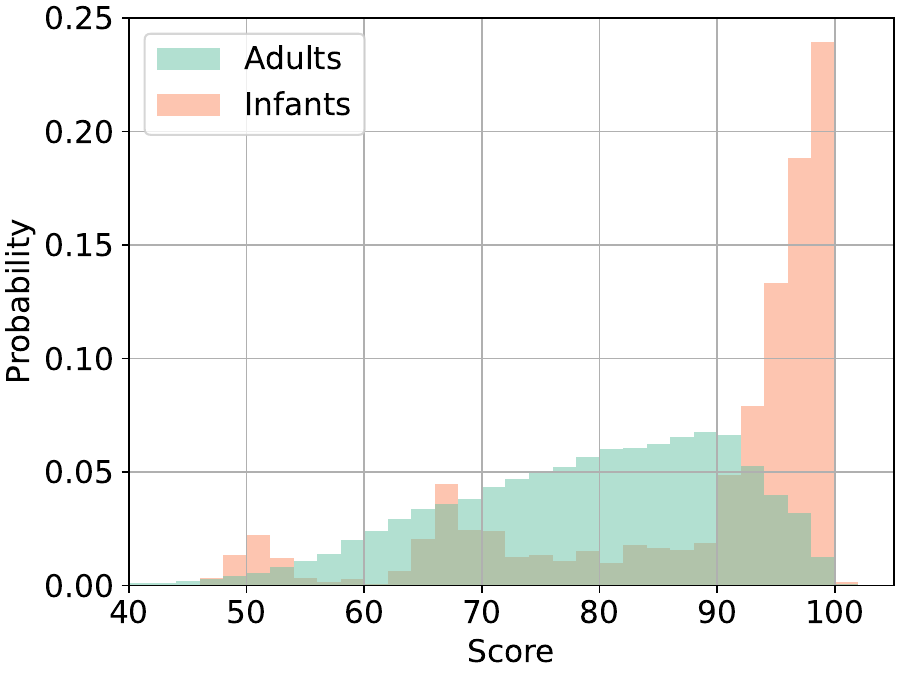}
        \caption{\texttt{USABLE\_IRIS\_AREA}}
        \label{fig:overall_USABLE_IRIS_AREA}
    \end{subfigure}\hfill
    \begin{subfigure}{0.25\linewidth}
        \includegraphics[width=\linewidth]{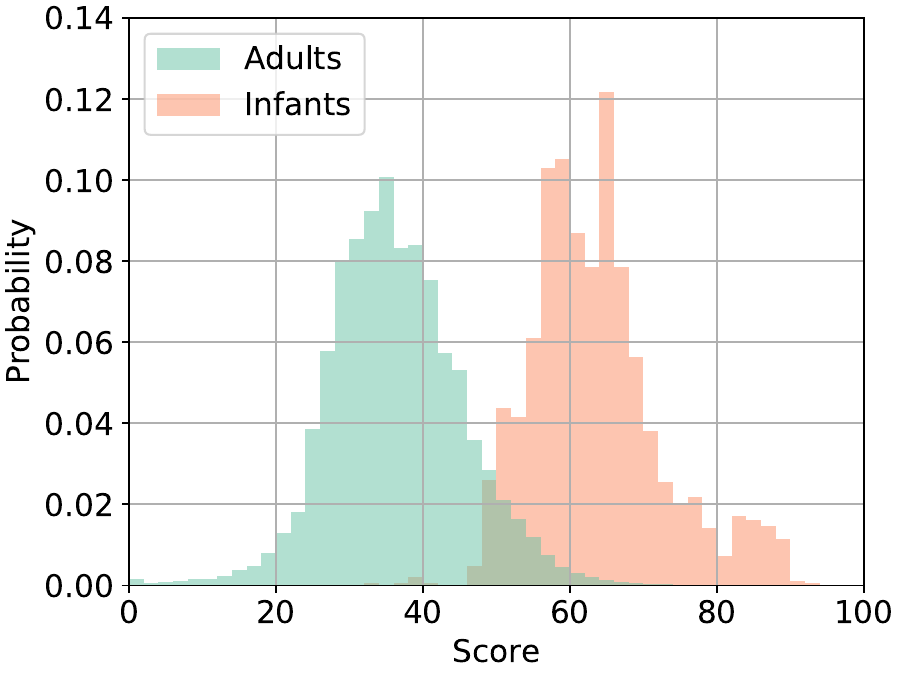}
        \caption{\texttt{IRIS\_SCLERA\_CONTRAST}}
        \label{fig:overall_IRIS_SCLERA_CONTRAST}
    \end{subfigure}\hfill
    \begin{subfigure}{0.25\linewidth}
        \includegraphics[width=\linewidth]{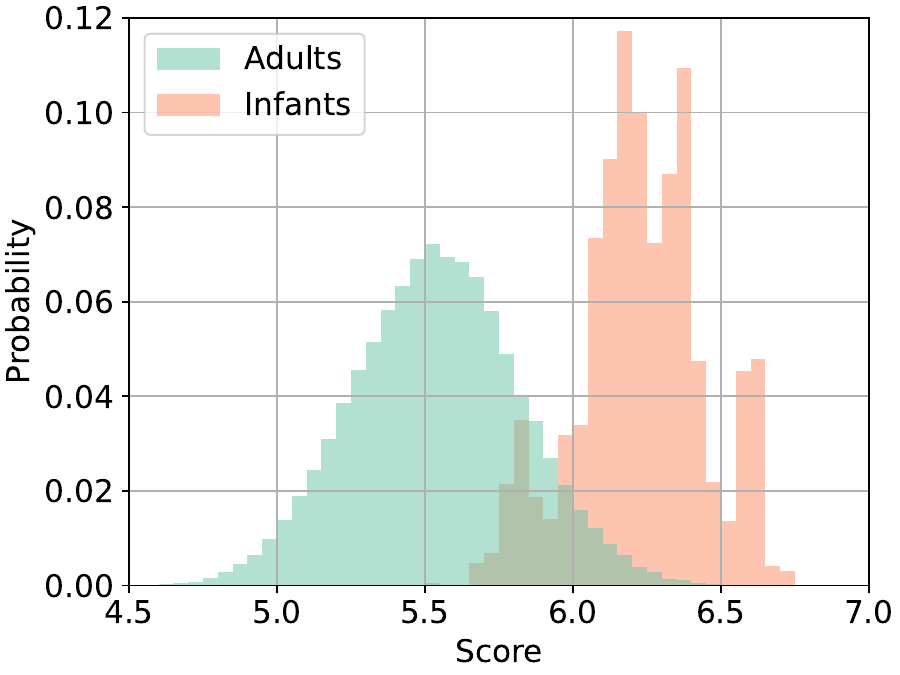}
        \caption{\texttt{GREY\_SCALE\_UTILISATION}}
        \label{fig:overall_GREY_SCALE_UTILISATION}
    \end{subfigure}\hfill
    \begin{subfigure}{0.25\linewidth}
        \includegraphics[width=\linewidth]{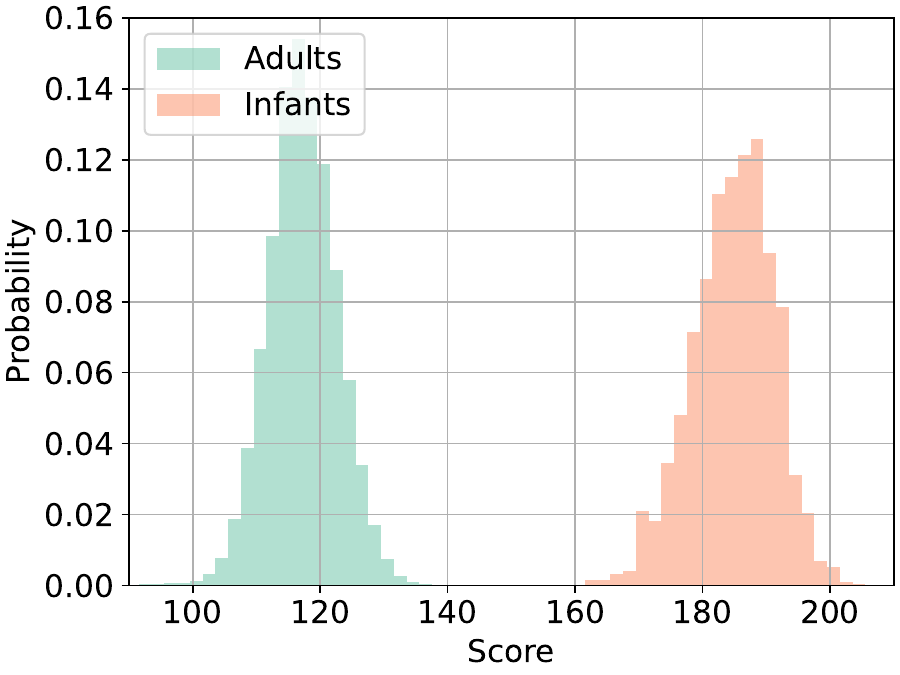}
        \caption{\texttt{IRIS\_RADIUS}}
        \label{fig:overall_IRIS_RADIUS}
    \end{subfigure}
    
    \begin{subfigure}{0.25\linewidth}
        \includegraphics[width=\linewidth]{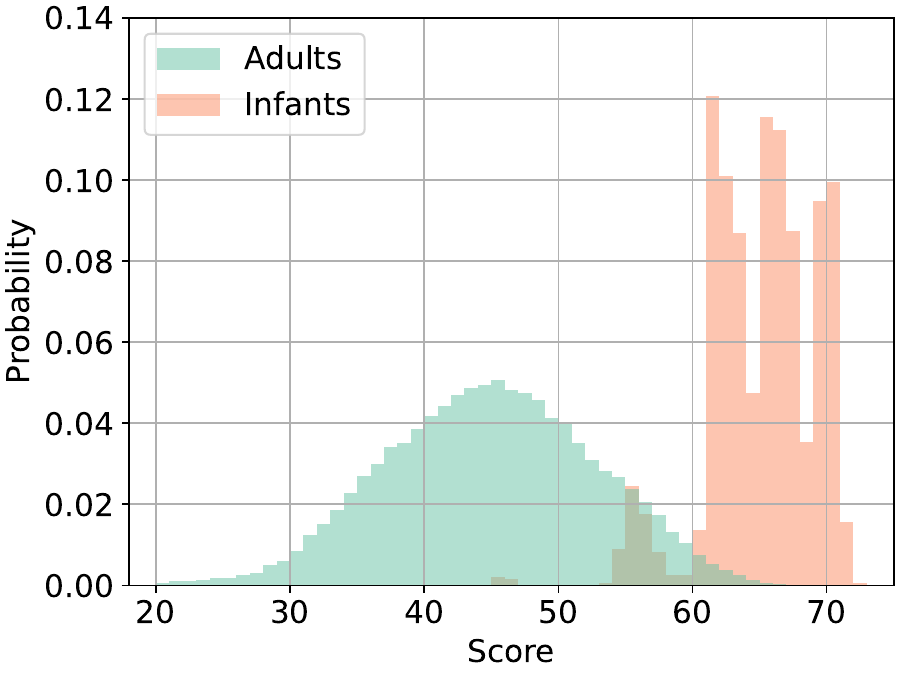}
        \caption{\texttt{PUPIL\_IRIS\_RATIO}}
        \label{fig:overall_PUPIL_IRIS_RATIO}
    \end{subfigure}\hfill
    \begin{subfigure}{0.25\linewidth}
        \includegraphics[width=\linewidth]{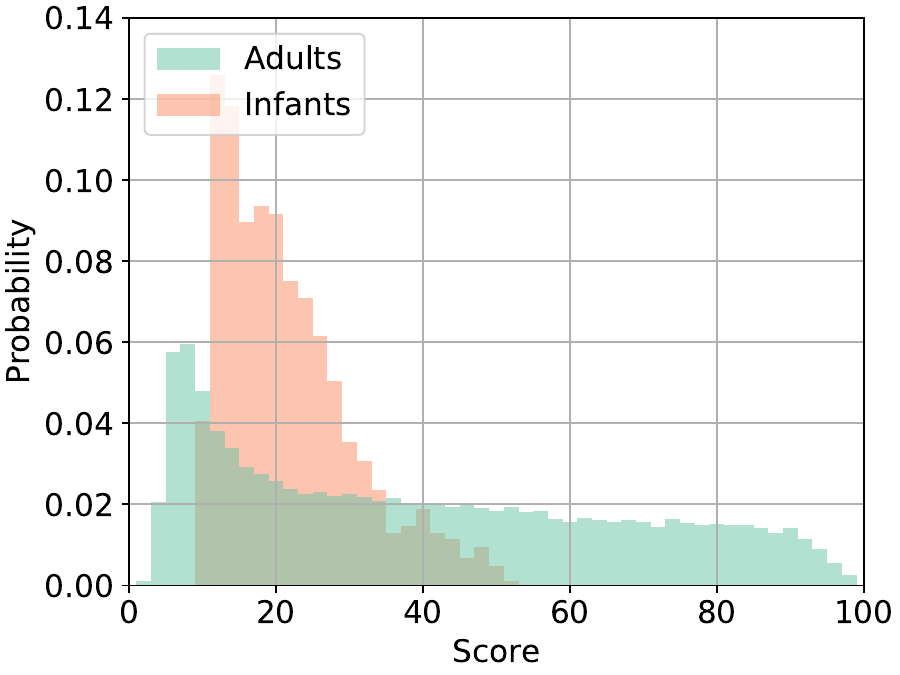}
        \caption{\texttt{SHARPNESS}}
        \label{fig:overall_SHARPNESS}
    \end{subfigure}\hfill
    \begin{subfigure}{0.25\linewidth}
        \includegraphics[width=\linewidth]{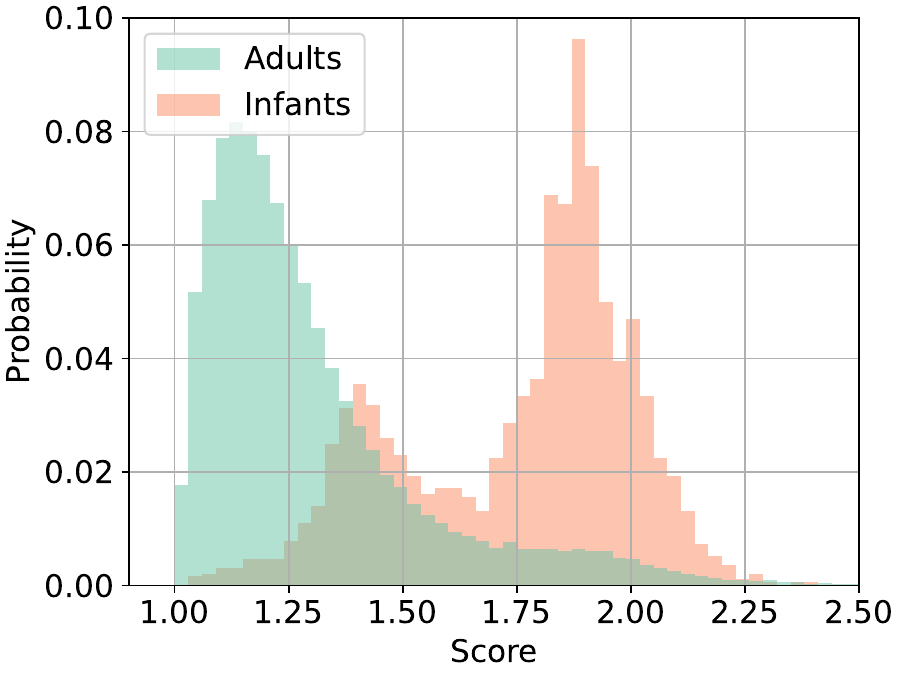}
        \caption{\texttt{MOTION\_BLUR}}
        \label{fig:overall_MOTION_BLUR}
    \end{subfigure}\hfill
    \begin{subfigure}{0.25\linewidth}
        \includegraphics[width=\linewidth]{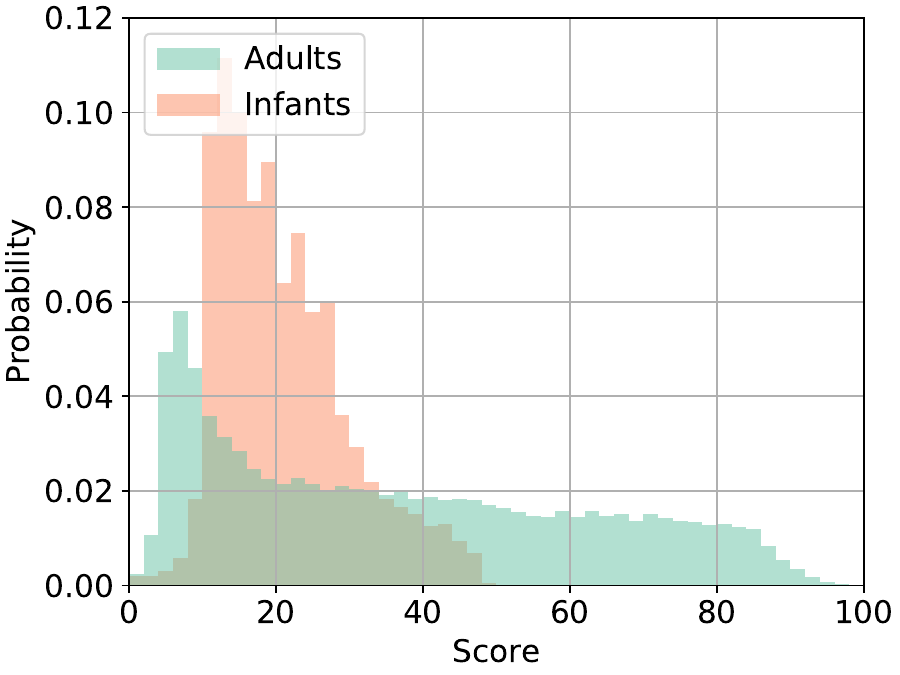}
        \caption{\texttt{OVERALL\_QUALITY}}
        \label{fig:overall_OVERALL_QUALITY}
    \end{subfigure}
    
    \caption{Distributions of the selected ISO/IEC 29794-6 iris image quality metrics calculated for adult and infant iris samples.}
    \label{fig:iso-score-dist}
\end{figure*}




As shown in Fig. \ref{fig:iso-score-dist}, 
infant iris images exhibit higher scores of the \verb+USABLE_IRIS_AREA+, \verb+IRIS_SCLERA_CONTRAST+, \verb+GREY_SCALE_UTILISATION+, \verb+IRIS_RADIUS+, \verb+PUPIL_IRIS_RATIO+, and \verb+MOTION_BLUR+ metrics compared to those obtained for adult samples. However, the \verb+SHARPNESS+ scores for infant samples range between 10 and 50, which is low compared to adult irises, possibly due to more challenging acquisition in healthcare setup, resulting in a higher probability of out-of-focus acquisition. Consequently, the \verb+OVERALL_QUALITY+ score for infant images is also lower than that of adult samples. One contributing factor could be the low \verb+SHARPNESS+ score and the \verb+IRIS_PUPIL_CONTRAST+ score, which is zero for all infant samples due to pupils being brighter than iris.

\subsection{Feasibility of Infant Iris Recognition}

A recent study suggested that while iris biometrics are effective for children aged two and older, their effectiveness diminishes for those under two years old \cite{nelufule2023infant}. To validate this claim 
we performed iris matching and calculated selected performance metrics including the decidability score $d'$ \cite{daugman2007new}, Failure-to-Match (FTM) rate, Equal Error Rate (EER), and Area Under the Receiver Operating Characteristics curve (AUC).

\begin{figure}[httb!] 
    \centering
    \begin{subfigure}[b]{0.5\linewidth}
        \centering
        \includegraphics[width=\linewidth]{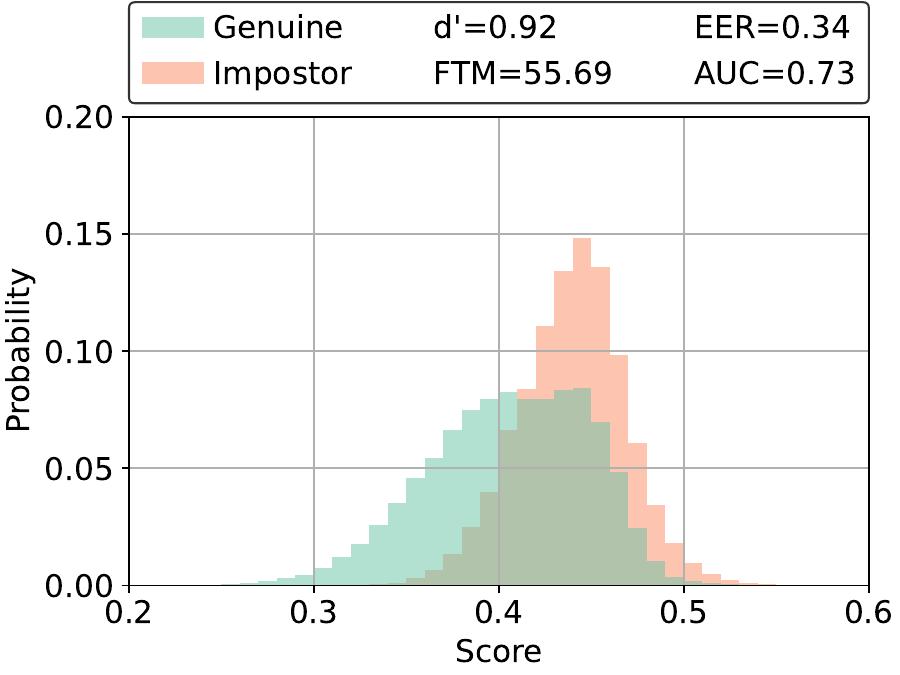}
        \caption{Vanilla HDBIF}
        \label{score-dist:hdbif-vanila}
    \end{subfigure}%
    \begin{subfigure}[b]{0.5\linewidth}
        \centering
        \includegraphics[width=\linewidth]{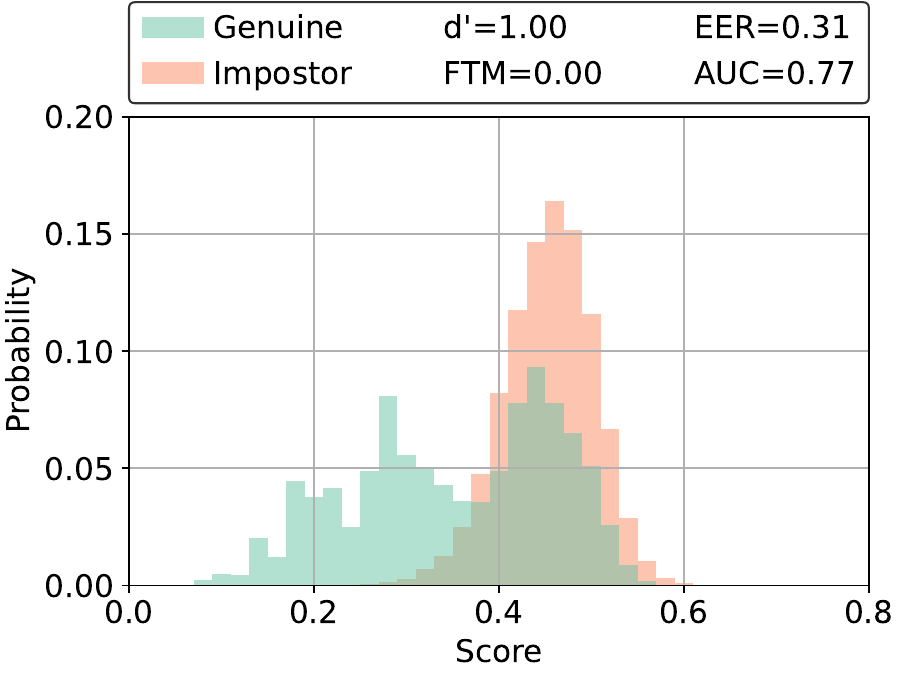}
        \caption{Vanilla OSIRIS}
        \label{score-dist:osiris-vanila}
    \end{subfigure}%
    \vskip3mm
    \begin{subfigure}[b]{0.5\linewidth}
        \centering
        \includegraphics[width=\linewidth]{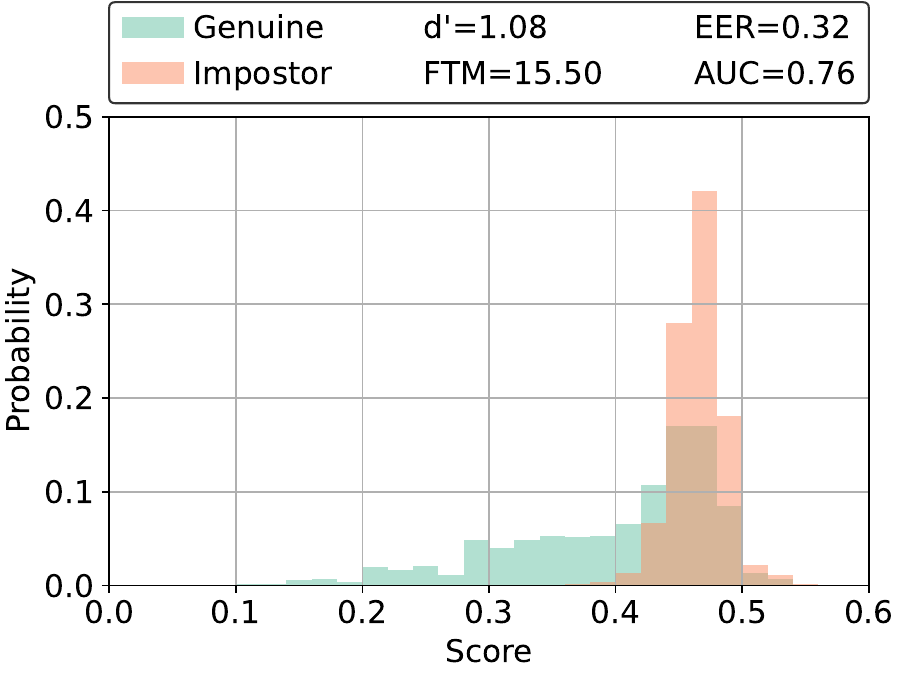}
        \caption{Vanilla USIT}
        \label{score-dist:usit-vanila}
    \end{subfigure}%
    \begin{subfigure}[b]{0.5\linewidth}
        \centering
        \includegraphics[width=\linewidth]{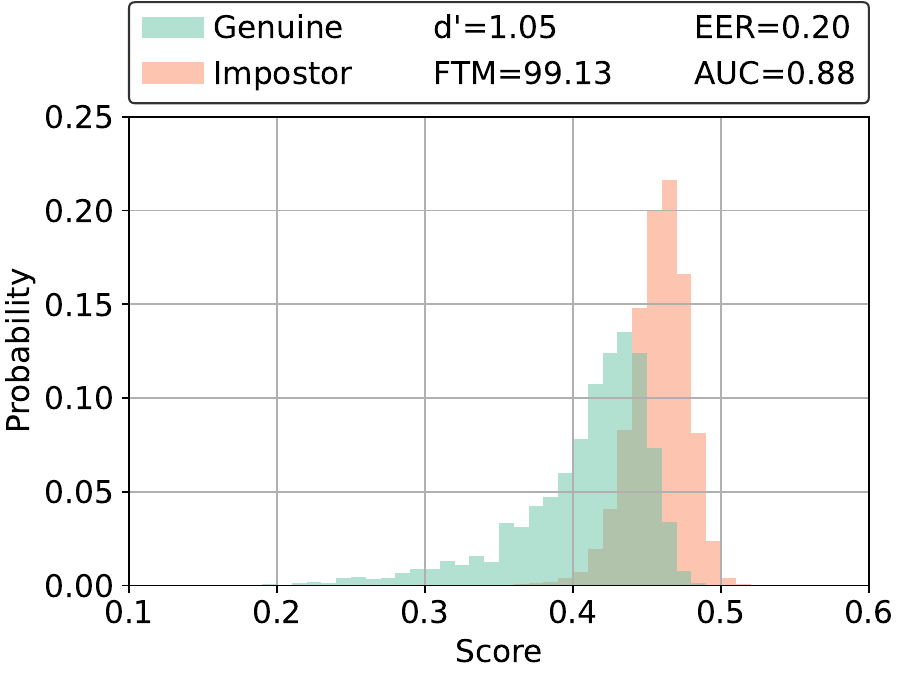}
        \caption{Vanilla WIRIS}
        \label{score-dist:wiris-vanila}
    \end{subfigure}%
    \caption{Distributions of genuine and impostor scores for four iris recognition methods that were able to process infant iris images. Vanilla segmentation models and encoding approaches were used, with default parameters suggested by the original method authors. Selected performance metrics ($d'$ statistic, Equal Error Rate (EER), Failure-to-Match rate and Area Under ROC curve (AUC) are also shown.}
    \label{fig:score-distribution}
\end{figure}

Fig. \ref{fig:score-distribution} presents the genuine and impostor score distributions obtained for vanilla iris recognition methods designed for adults. All possible genuine and impostor pairs, possible to be generated from 1,920 infant images, were considered in this study. VeriEye is not shown in Fig. \ref{fig:score-distribution} since this commercial matcher was not able to process any infant iris image (thus FTM=100\% for this method). Also, DGR is not shown in Fig. \ref{fig:score-distribution} since this approach proposes only texture encoding but not iris image segmentation. The FTM rates for USIT and OSIRIS vanilla methods were 15.50\% and 0.0\%, respectively, while HDBIF and WIRIS approaches exhibited much higher FTM rates of 55.69\% and 99.13\%, respectively, reflecting difficulties due to insufficient usable iris bits available after segmentation stage.


\begin{figure}[htb!] 
    \centering
    \begin{subfigure}[b]{0.5\linewidth}
        \centering
        \includegraphics[width=\linewidth]{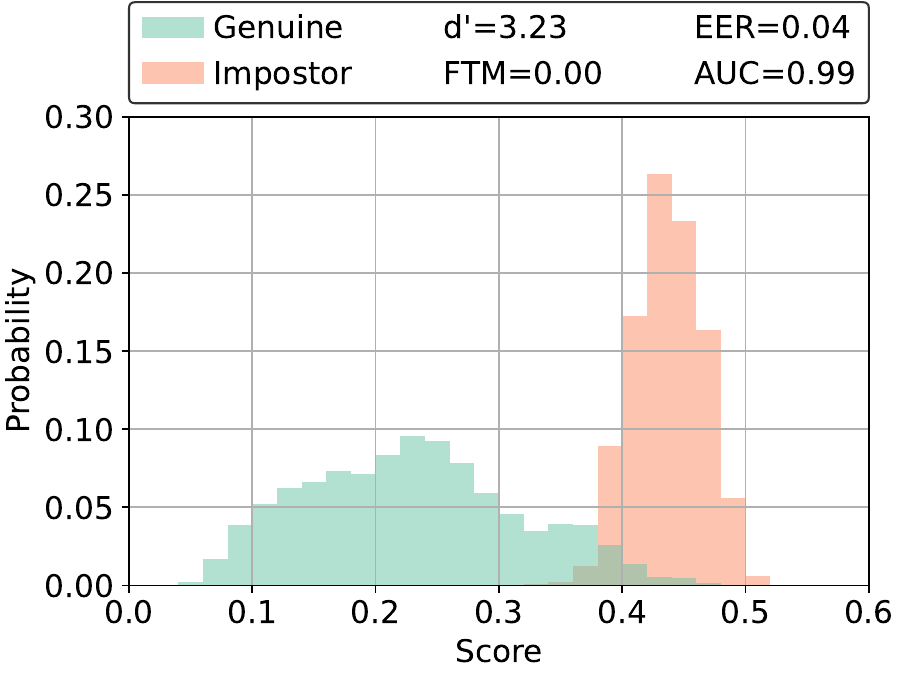}
        \caption{IIS + HDBIF encoding}
        \label{score-dist:hdbif-norm}
    \end{subfigure}%
    \begin{subfigure}[b]{0.5\linewidth}
        \centering
        \includegraphics[width=\linewidth]{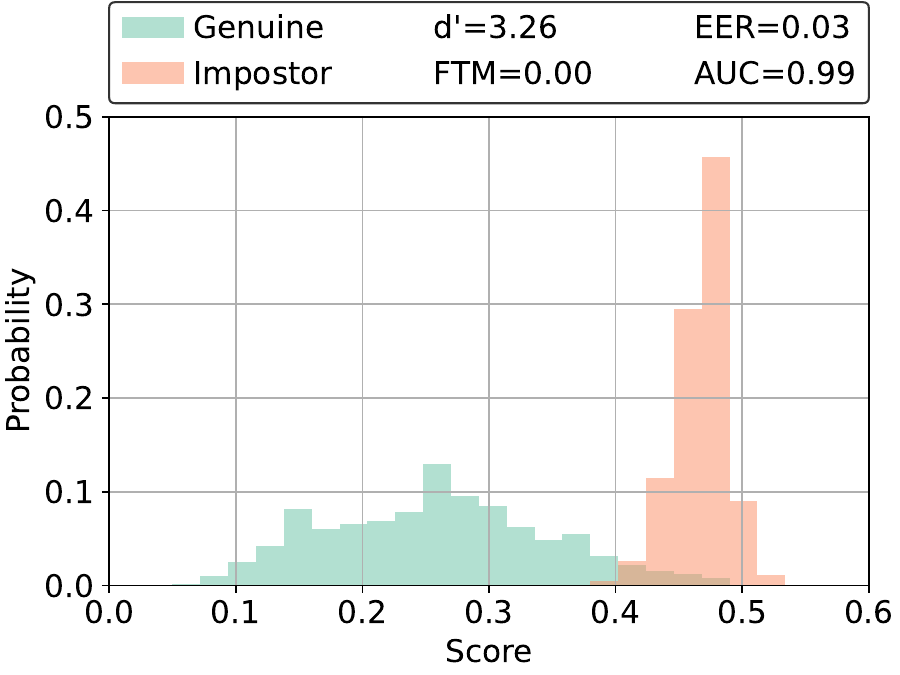}
        \caption{IIS + OSIRIS encoding}
        \label{score-dist:osiris-norm}
    \end{subfigure}%
    \vskip3mm
    \begin{subfigure}[b]{0.5\linewidth}
        \centering
        \includegraphics[width=\linewidth]{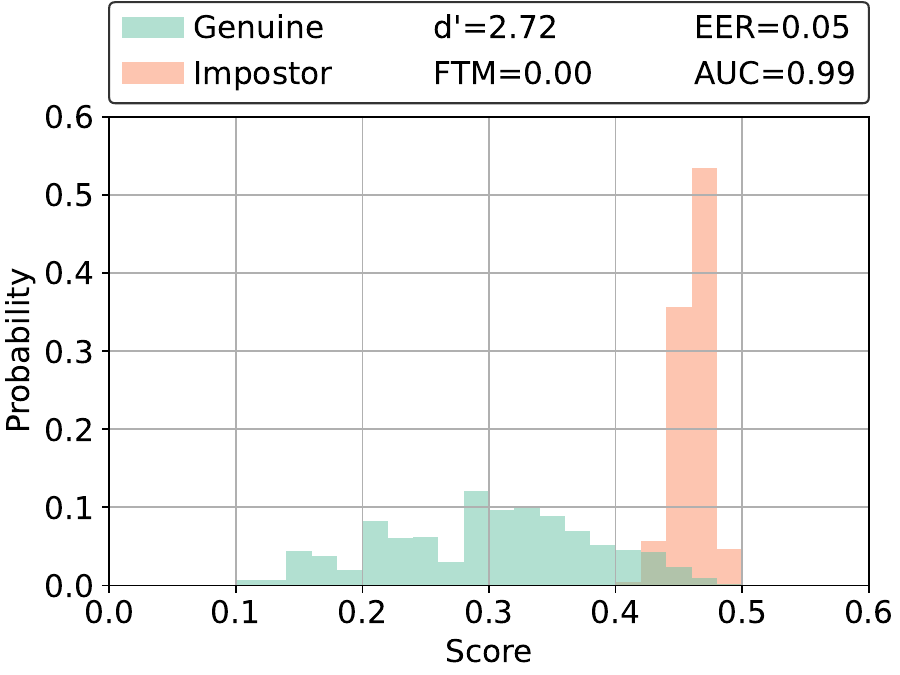}
        \caption{IIS + USIT encoding}
        \label{score-dist:usit-norm}
    \end{subfigure}%
    \begin{subfigure}[b]{0.5\linewidth}
        \centering
        \includegraphics[width=\linewidth]{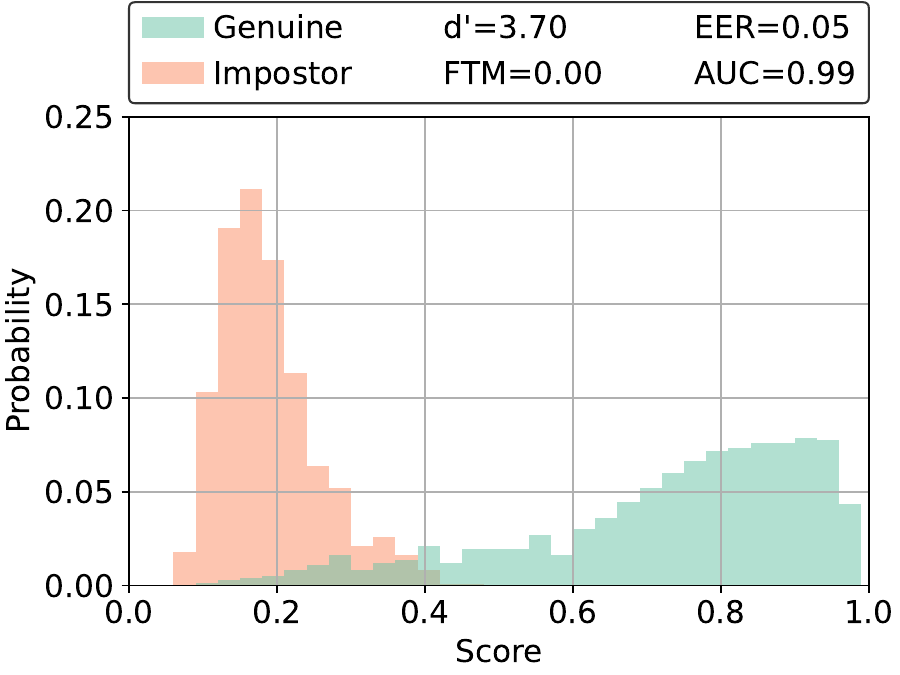}
        \caption{IIS + DGR encoding}
        \label{score-dist:dgr-norm}
    \end{subfigure}%
    
    \caption{Distributions of genuine and impostor scores for four combinations of the proposed Infant Iris Segmentation (IIS) model and various iris texture encoding methods originating from four different iris recognition approaches. We see a significant boost in performance when the proposed segmentation model is incorporated into infant iris recognition pipeline, compared to Fig. \ref{fig:score-distribution}.}
    \label{fig:score-distribution-with-new-segmentation} 
\end{figure}

Fig. \ref{fig:score-distribution-with-new-segmentation} presents results after replacing vanilla segmenters with the proposed infant iris segmentation (IIS) model. The results clearly indicate a substantial improvement in performance across all matchers. 
%
Especially, the FTM rate was reduced to 0.0\% across the board.
The HDBIF matcher saw a remarkable improvement, with the decidability score ($d'$) rising from 0.92 to 3.23, EER dropping from 34\% to 4\%, and AUC increasing from 0.73 to 0.99. Similar performance gains are also observed for the USIT and OSIRIS encoding approaches. Notably, the DGR iris texture encoder allowed to achieve the highest $d'$ score of 3.70. 
The OSIRIS matcher integrated with our segmentation model demonstrated the best performance, achieving an EER of 3\%. 
These promising results obtained after replacing the standard iris segmentation with the proposed segmenter suggest that there are chances of building an effective infant iris recognition method with infant-specific iris image processing.

\section{Discussion}

\paragraph{Infant iris recognition performance} Our study aimed to validate whether an effective iris recognition system can be built for infants. While the VeryEye matcher was unable to process newborn images due to its stringent  requirements tied to properties of adult iris images, other matchers, particularly when integrated with our proposed segmentation model, demonstrated significantly improved performance, with EER not exceeding 4\% and $d'$ score ranging from 2.7 to 3.7, what indicates a reasonable capability in distinguishing between genuine and impostor infant iris images. These results suggest that with appropriate preprocessing and segmentation, iris recognition can become a new and reliable identification means for infants and neonates.

\paragraph{Limitations} This study for the first time, to our knowledge, proposes and evaluates an iris recognition system specifically designed for infants, with two custom-designed and crucial components: hardware (acquisition) and image processing (segmentation). Being an early-stage study it also has several limitations, that we want to list here and address in immediate future research efforts. First, all infant iris images were acquired in a single session. One of the most crucial questions, after seeing in this paper good chances of having such images properly processed and encoded, is how stable the iris pattern is in the first weeks and months of human's life. Second limitation is a relatively small number of subjects in our dataset, which includes only 1,920 samples from 17 individuals, all collected in the same hospital. This limited sample size may not capture the variability present in larger, more diverse populations. Furthermore, the absence of publicly available datasets of infant irises, or any other iris recognition methods specifically designed for infant iris recognition restrict our ability to benchmark our results against broader external data sources and other algorithms for infant iris recognition.

We do hope, however, that this pioneering study will contribute to expanding the iris recognition methodology to infants and neonates, and will bring identification means increasing safety of the youngest members of our population. The proposed segmentation model, source codes and synthesized infant iris images are made available along with this paper to facilitate achieving this goal.

\paragraph{Acknowledgments} This material is based upon work partially supported by the National Science Foundation under Grant No. 2237880. Any opinions, findings, and conclusions or recommendations expressed in this material are those of the authors and do not necessarily reflect the views of the National Science Foundation.


\balance
{\small
\bibliographystyle{ieee_fullname}
\bibliography{egbib}
}

\end{document}